\RequirePackage{fix-cm}
 \documentclass[smallextended]{svjour3}       
\smartqed  
\usepackage{float}
\usepackage{amsmath}
\usepackage{graphicx}
\usepackage{subimages}
\graphicspath{{img/}}
\usepackage[caption=false]{subfig}
\usepackage{xcolor}
\usepackage{multirow}

\newcommand{\secref}[1]{Section~\ref{sec:#1}}
\newcommand{\secrefs}[2]{Sections~\ref{sec:#1}~and~\ref{sec:#2}}
\renewcommand{\eqref}[1]{Equation~(\ref{eq:#1})}
\renewcommand{\figref}[1]{\figurename~\ref{fig:#1}}

\newcommand{\etalcite}[1]{{\emph{et al.}$\!$}~\cite{#1}}
\newcommand{\img}{\ensuremath{{\mathcal{I}}}}
\newcommand{\quantizedimg}{\ensuremath{\tilde{\mathcal{I}}}}

\newcommand{\histogram}{\ensuremath{\mathcal{H}_{ij}}}
\newcommand{\bgij}{\ensuremath{\mathcal{B}_{ij}}}

\setfigdir{./images}
 \journalname{}

\begin{document}

\title{LRCN-RetailNet: A recurrent neural network architecture for accurate people counting 
}


\author
{
    Lucas Massa \and Adriano Barbosa \and Krerley Oliveira \and Thales Vieira 
}


\institute{Lucas Massa \at
              Institute of Computing - Federal University of Alagoas, Maceió, AL, Brazil\\
              \email{lmm@ic.ufal.br} 
           \and
           Adriano Barbosa \at
           Faculty of Exact Sciences and Technology - Federal University of Grande Dourados, Dourados, MS, Brazil\\
              \email{adrianobarbosa@ufgd.edu.br} 
              \and
           Krerley Oliveira \at
               Institute of Mathematics - Federal University of Alagoas, Maceió, AL, Brazil\\        
              \email{krerley@im.ufal.br} 
              \and
           Thales Vieira \at
            Institute of Computing - Federal University of Alagoas, Maceió, AL, Brazil\\
              \email{thales@ic.ufal.br} 
}

\date{Received: date / Accepted: date}

\maketitle

\begin{abstract}
Measuring and analyzing the flow of customers in retail stores is essential for a retailer to better comprehend customers' behavior and support decision-making. Nevertheless, not much attention has been given to the development of novel technologies for automatic people counting. We introduce LRCN-RetailNet: a recurrent neural network architecture capable of learning a non-linear regression model and accurately predicting the people count from videos captured by low-cost surveillance cameras. 
The input video format follows the recently proposed RGBP image format, which is comprised of color and people (foreground) information. 
Our architecture is capable of considering two relevant aspects: spatial features extracted through convolutional layers from the RGBP images; and the temporal coherence of the problem, which is exploited by recurrent layers. We show that, through a supervised learning approach, the trained models are capable of predicting the people count with high accuracy. 
Additionally, we present and demonstrate that a straightforward modification of the methodology is effective to exclude salespeople from the people count.
Comprehensive experiments were conducted to validate, evaluate and compare the proposed architecture. Results corroborated that LRCN-RetailNet remarkably outperforms both the previous RetailNet architecture, which was limited to evaluating a single image per iteration; and a state-of-the-art neural network for object detection. Finally, computational performance experiments confirmed that the entire methodology is effective to estimate people count in real-time.
\end{abstract}

\section{Introduction}
\label{intro}

Customer behavior analysis is an essential task to gain relevant insights and drive the decision-making process of retailers. As a result, a merchant may enhance customer experience, optimize operational costs and store performance, and consequently maximize profitability~\cite{liu2017customer,hawkins2010consumer}. 

In particular, an inadequate sales staff scheduling may result both in low rates of conversion, \emph{i.e.} transforming a browsing customer into a buying customer; and in sub-optimal labor costs, which is widely known to be a relevant expenditure in the budget of a retail store~\cite{lam1998retail}.
Although buying behaviors can be easily tracked through transaction logs, it is not straightforward to comprehend more hidden patterns such as the flow of non-buyers, since they are usually harder to track.  

Fortunately, surveillance cameras have become ubiquitous in public indoor environments and particularly in retail stores. Such cameras provide images that may be exploited to identify and track any visible individual. 

While it is impractical for a human to analyze several hours of video every day, recent advances in Deep Learning algorithms and hardware have allowed the development of extremely robust Computer Vision tools that may replace a human in tasks like real-time image classification and object localization~\cite{Redmon_2016_CVPR}. In particular, it is now feasible to investigate and develop effective algorithms to automatically extract people flow information from affordable cameras, although it is still a scarcely studied problem~\cite{retailnet}.

In this paper we focus on the specific problem of accurately counting the number of people inside a store in real-time, using data collected from a low-cost surveillance camera. We introduce LRCN-RetailNet: a recurrent neural network architecture for accurate people counting that exploits the spatio-temporal coherence of videos acquired from surveillance cameras.  
It is worth mentioning that, to the best of our knowledge, this particular problem has not been much studied in the literature. Recently, Nogueira~\etalcite{retailnet} proposed an approach that combines a foreground detection method with a CNN regression model called RetailNet to predict the people count from single images. Our proposed approach resembles that previous work pipeline, but taking advantage of the spatio-temporal coherence of the problem: LRCN-RetailNet combines essential qualities of both the original RetailNet and the Long-term Recurrent Convolutional Networks architecture (LRCN, \cite{donahue2015long}). This synergy allows the resulting network to exploit compositional representations in space and time, which are mainly related to the density of people in the spatial domain of the images.

The proposed LRCN-RetailNet architecture is trained through supervised learning. We make use of low-resolution RGB videos acquired from a single low-cost surveillance camera and then annotated by a human trainer according to the people count of each image.
The videos are converted to the RGBP image representation~\cite{retailnet} by extracting its foreground (or people) channel using an algorithm developed to deal with the peculiarities of a retail store.
Small videos (or sequences of RGBP images) are considered as input to the LRCN-RetailNet, which is essentially a deep regression model trained to predict the people count in the last frame of the video.
We emphasize that, in the studied setting, people in a store are prone to be partially occluded by furniture, objects and/or other people. Furthermore, customers often appear in extreme poses, such as when seated to try on shoes. However, we claim that such issues are mitigated when several images from a video are jointly exploited by LRCN-RetailNet. Additionally, we present an adaptation of the proposed methodology to make the network count only the visible customers in the images (disregarding salespeople), since the flow of customers is more relevant to customer behavior analysis.

We show in our experiments that the proposed recurrent architecture significantly outperforms both the original RetailNet and the state-of-the-art object detection method YOLO v3~\cite{yolov3} in the people counting task. We also conducted comprehensive experiments to validate the whole approach (including customer-only count); optimize hyper-parameters; evaluate transfer learning strategies; and measure the computational performance to confirm that LRCN-RetailNet is indeed capable of performing real-time people counting.



\section{Related work}
The specific problem of accurate people counting has not been widely studied in the literature. Nevertheless, there is already a large body of literature on crowd density estimation methods, which are closely related to our work. 
According to Sindagi~\etalcite{sindagi2018survey}, such methods can be grouped into three main categories: detection-based, regression-based and density estimation-based methods. In addition to discussing each of these three categories, we also mention a few methods based on depth images (RGBD) and give special attention to deep learning methods, since they are more closely related to our proposed method. We refer the reader to surveys on crowd counting methods \cite{loy2013crowd,sindagi2018survey} for a more comprehensive presentation.

Detection-based methods rely on the recognition of each visible person in the image and make use of hand-crafted features such as Haar wavelets~\cite{viola2004haar}, Histogram of Oriented Gradients (HOG, ~\cite{dalal2005histograms}), edgelet features~\cite{wu2005edgelet} and shapelet features~\cite{sabzmeydani2007detecting}.
Despite achieving good results in low-density crowd scenes, this class of techniques is unreliable in scenes containing many partially occluded objects and high-density crowds. 
Classifiers for specific body parts built on top of part-based detection~\cite{felzenszwalb2009object,li2008partbased} and shape-based~\cite{zhao2008shapelearning} methods addressed this issue.
Although better results can be achieved with part-based detection, it is still challenging to recognize people in low-resolution images acquired from affordable surveillance cameras, and even harder when individuals are strongly occluded. 

Regression-based methods achieve better results in scenes with dense crowds and background clutter by using global and local features like foreground features extracted by background subtraction~\cite{bouwmans2014traditional}, blob-based holistic features such as area, perimeter and edges~\cite{chan2008privacy}, and gray level co-occurrence matrices (GLCM)~\cite{Chen_featuremining}. Such features were proposed to be combined with different regression techniques including linear regression~\cite{Paragios2001AMA}, ridge regression~\cite{Chen_featuremining} and neural networks~\cite{marana1998tacm}.

Density estimation-based techniques employ linear~\cite{lempitsky2010learningtocount} or non-linear~\cite{Pham_2015_ICCV} mappings between local patch features and density maps. In \cite{wang2016objectcounting}, a density estimation approach based on subspace learning enables the learning of an embedding of each subspace formed by image patches. Xu~\etalcite{xu2016crowd} pointed out that such crowd density methods use only a small set of features due to the computationally expensive Gaussian and Ridge regressions. They proposed to use a richer set of features combined with random forest regression to enhance the results and reach the necessary computational efficiency and scalability.

RGBD image and videos help to mitigate the issues caused by clutter and occlusions by merging depth and color information.
Simplified Local Ternary Patterns (SLTP, \cite{rauter2013reliable}) were developed to detect and track the human head from depth images in a top-view camera system. The so-called water filling method~\cite{zhang2012water} is an unsupervised method that uses a vertical Kinect sensor and finds the regions of the depth image with a suitable local minimum.
A few methods make use of the combination of RGB images and depth information for people counting~\cite{sun2019benchmark}. Gao~\etalcite{gao2016people} combines a water filling method with an SVM classifier using HOG features to detect head candidates, and track the trajectories from depth videos. Another approach was proposed by Liu~\etalcite{liu2013real}, which trains an SVM classifier to detect the upper part of a human body. Although depth information is valuable to perform people counting, it is not practical for retail stores and indoor environments in general, since the depth acquisition technologies impose depth range limits. Also, their affordability and availability are still not comparable to the ubiquitousness of surveillance cameras. Similarly to these techniques, we also consider valuable information in addition to the color image, but in the form of a foreground channel.

Deep learning methods have also been applied to the crowd counting problem and the results have been very promising~\cite{sindagi2018survey}.
Liu~\etalcite{liu2017passenger} proposed a counting system for public transportation based on the combination of convolutional neural networks (CNN), which detects the passengers, and Spatio-Temporal Context (STC)~\cite{zhang2014fast}, to track passengers heads.
Wei~\etalcite{wei2019boosting} recently proposed an approach to crowd counting that combines support vector regression and spatial-temporal multi-features by joining super-pixel based multi-appearance features and multi-motion features, and then combine with a deep attribute learning architecture based on the VGG16 model.
Zhang~\etalcite{Zhang_2016_CVPR} propose a multi-column convolutional neural network (MCNN) architecture to map the image to the corresponding crowd density map. The features learned by each CNN column are adaptive to variations in people (or head) size and the density map is computed based on geometry-adaptive kernels.
Boominathan~\etalcite{boominathan2016crowdnet} proposed the CrowdNet framework for estimating crowd density from images of highly dense crowds. CrowdNet uses a combination of deep and shallow architecture which can capture both high-level semantic information (face detectors) and low-level features (blob detectors). Such methods, however, aim to roughly estimate the number of people in crowded outdoor images, and consequently cannot be directly applied to accurately monitor the people flow in an indoor environment. 

In this direction, RetailNet~\cite{retailnet} was recently proposed to perform accurate people counting in retail stores using only color images acquired from an inexpensive surveillance camera. By first extracting the background in an appropriate manner to identify visible people, an RGBP representation is proposed to train a CNN regression model, achieving robust results in real-world situations. However, the spatio-temporal coherence of the problem is not exploited: single images are separately evaluated by the CNN.
We propose a pipeline similar to \cite{retailnet}, but taking advantage of the spatio-temporal coherence by evaluating a sequence of several images. We claim that, compared to RetailNet, the proposed recurrent architecture may better deal with challenges like high occlusion and extreme human poses, due to the larger amount of data considered for prediction.


\section{Methodology}
\label{sec:methodology}
\begin{figure*}[!t]
\includegraphics[width=\linewidth]{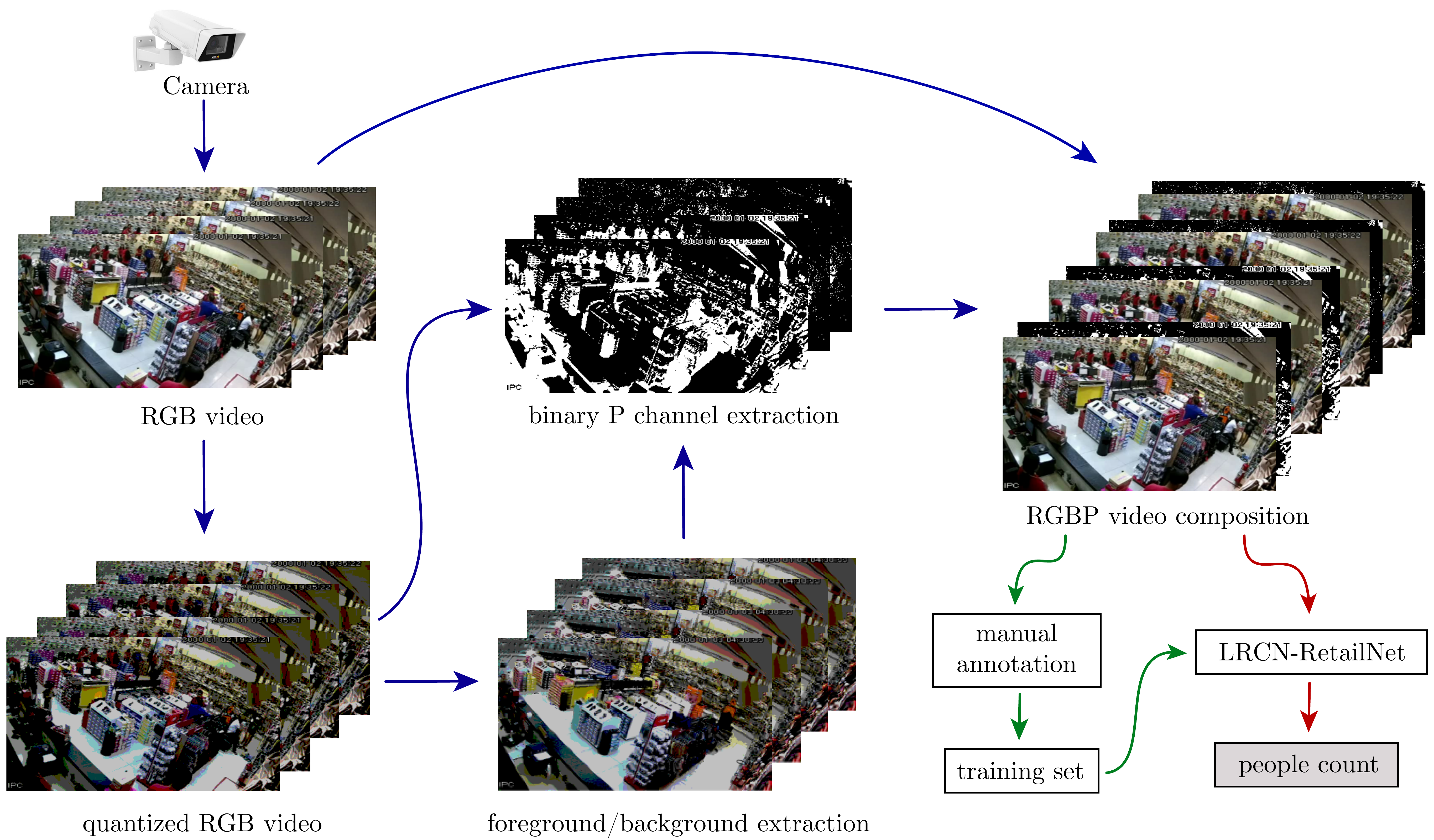}
\caption{Overview of our approach: a preprocessing step (blue arrows flow) is first applied to the images before following to compose the training set of the neural network (green arrows flow) or to predict the people count (red arrows flow). In the preprocessing step, images from RGB video acquired from a low resolution surveillance camera are first quantized, and then used to initialize or update a P image representing the foreground (or people). The P images are merged to the original RGB images to compose RGBP images. In the training phase, RGBP videos are annotated by a human trainer before composing a training set used to train the LRCN-RetailNet. In the prediction phase, the trained model may be employed in real-time to accurately predict the people count in the last frame of an RGBP video given as input.}
\label{fig:overview}
\end{figure*}

Our methodology follows a pipeline that resembles the one proposed by Nogueira~\etalcite{retailnet}, which is also intended to perform accurate people counting from color images. \figref{overview} gives an overview of our approach, which is comprised of three main parts: a preprocessing step, a training step, and a prediction step. We consider that only low-resolution RGB video captured from affordable surveillance cameras are given as input.

The supervised learning algorithm is based on a recurrent convolutional architecture derived from the well known \emph{Long-term Recurrent Convolutional Networks} (LRCN, \cite{donahue2015long}), which have been successfully applied to many video recognition tasks~\cite{Amaral2019}. 
We consider the people counting task as a regression problem whose input is a sequence of RGBP images, here called RGBP video. This representation is derived from the recently proposed RGBP image, which combines color and foreground (people) information~\cite{retailnet}.
RGBP images and videos are computed in the preprocessing step by considering the spatio-temporal coherence of the problem to detect and extract foreground images. This is accomplished by analyzing the most recent RGB images acquired from the camera. The RGBP videos are then utilized in the other parts of the method: training and prediction. 

In the training phase, RGB videos of the retail store are first collected, converted to the RGBP format, and then annotated with the number of people. Here, it is worth mentioning that, differently from \cite{retailnet}, the training examples are sequences of $T$ continuous images in time (instead of single images) annotated according to the number of people in the last image of the sequence. The resulting supervised training set is used to train an LRCN-RetailNet regression model. 

In the prediction step, the people count of a queried image $\mathcal{I}$ is estimated by considering the sequence (or video) $\mathcal{S}$ of $T$ RGB images comprised of $\mathcal{I}$ and its  $T-1$ preceding RGB images in time.
The same preprocessing is applied to convert $\mathcal{S}$ to the RGBP video format, which is then given as input to the trained regression model. Finally, the people count of the last image of $\mathcal{S}$ is predicted by the model. It is worth mentioning that the prediction step can be employed in real-time to promptly estimate the people flow in retail-stores. In the following subsections, we describe each of the aforementioned steps.

\subsection{Foreground/background detection}
\label{sec:foreback-detection}
Foreground/background detection is a recurrent problem for several Computer Vision applications and has been investigated for many years~\cite{bouwmans2014traditional}. In the particular context of retail stores, it is not appropriate to adopt simple approaches such as using a single static image as a reference for the background due to several challenges inherent to this problem:
\begin{enumerate}
    \item furniture and products are frequently changing positions;
    \item shadows and illumination varies in time;
    \item people often remain motionless for a period of time, particularly salespeople and customers browsing or trying on products, or in the cashier line.
\end{enumerate}
 Thus, to detect the background we adopt the same approach proposed by Nogueira~\etalcite{retailnet}, which comprises a preprocessing step followed by a proper background initialization and iterative dynamic background updates. In what follows, we describe each of these steps.

\paragraph{Image preprocessing.} Firstly, every captured image $\img$ is resampled to fit the spatial dimensions of the neural network input, which is set to $400\times 225$ pixels in this work. To mitigate illumination issues, resampled images are then uniformly quantized to $\lambda$ levels ($\lambda_c$ levels per channel). We follow \cite{retailnet} and set $\lambda_c=4$, resulting in quantized images $\quantizedimg$ with $\lambda=4^3=64$ different quantized levels.

\paragraph{Background initialization.} To generate the initial background $\mathcal{B}$, a circular image buffer $\mathcal{C}$ initially accumulates $\eta$ images sequentially sampled from a quantized RGB video, possibly in real-time. As in \cite{retailnet}, we fix $\eta$ to 100 and the sampling rate to one frame per second to accumulate information from $100$ consecutive seconds. Such values were sufficiently robust to remove salespeople from the background in practice. 
When the circular image buffer $\mathcal{C}$ becomes filled, histograms $\histogram$ are computed for each pixel $(i,j)$ of the spatial domain of the images in $\mathcal{C}$. Each histogram is comprised of $\lambda$ bins, where each bin represents the frequency in which a quantized level appears in the pixel position $(i,j)$ of the images in $\mathcal{C}$. More formally,
$$\histogram[l]= \#\left\{\quantizedimg \in \mathcal{C} \mid\quantizedimg_{ij}=l  \right\} \, ,\quad l=1 \dots \lambda .$$
 The background color of pixel $(i,j)$ is then defined as the most frequent color in that pixel, \emph{i.e.} the mode of pixel $(i,j)$:
$$\bgij = \mbox{mode}(\histogram) \, , \quad i=1\dots 225, \, j=1 \dots 400 \, .$$

\paragraph{Background updates.} Despite the high quality of the backgrounds provided by the aforementioned method, it is still necessary to regularly update it due to changes in both the position of visible objects in the background, such as shoes boxes or furniture; and in the illumination of the store. The update process from time step $t-1$ to $t$ make use of the previous background $\mathcal{B}^{t-1}$, the current histograms $\histogram^{t-1}$, and the circular buffer $\mathcal{C}$, which keeps being updated with the upcoming images acquired from the camera. Before overwriting the oldest image $\quantizedimg_o$ in $\mathcal{C}$ with a new image $\quantizedimg_n$, it is necessary to remove its data from the histograms $\histogram^{t-1}$ and update them with data from $\quantizedimg_n$. This computation results in the subsequent histograms $\histogram^{t}$, which are next utilized to update the previous background colors $\mathcal{B}_{ij}^{t-1}$ to updated background colors $\mathcal{B}_{ij}^{t}$. However, differently from the initialization strategy, here a more cautious condition is employed:
$$\bgij^t = \left\{
\begin{array}{cl}
    \mbox{mode}(\histogram^{t}), & \mbox{if } \max(\histogram^t) \ge \tau\cdot\eta \\
    \bgij^{t-1}, & \mbox{otherwise},
\end{array}
\right.$$
where $\tau$ is a threshold to ensure that the update occurs only if the samples in $\histogram$ are highly concentrated in a single bin. This additional constraint to change the previous background colors is imposed to avoid standstill people to be incorrectly recognized as background in many cases. As in \cite{retailnet}, we adopt $\tau=0.8$. 

\paragraph{Foreground detection:} Each background image $\mathcal{B}$ is used to compute a binary image P, representing the spatial distribution of people in the RGB image $\quantizedimg$. This is achieved by taking the absolute difference between $\quantizedimg$ and $\mathcal{B}$, and thresholding a grayscale version of such difference image, as summarized in the following equation:
$$P_{i,j} = \left\{
\begin{array}{cl}
    1, & \mbox{if } \mbox{gray}(|\quantizedimg_{ij}-\mathcal{B}_{ij}|) > \beta \\
    0, & \mbox{otherwise}
\end{array} , \quad \quad \forall i,j \, ,
\right.$$
where $\mbox{gray}(\cdot)$ is a grayscale conversion function and $\beta$ is a small binarization threshold set to $0.1$. Note that $P$ is expected to be $1$ in regions of the image occupied by people, thus providing higher level information to the neural network. Lastly, $P$ is merged to the original image $\img$ to constitute a four channel RGBP image, which is used as input to the proposed supervised learning approach.

\subsection{Supervised learning approach} \label{sec:supervisedlearning}
We consider the people counting problem as a regression analysis process. A supervised learning approach is employed to train a deep neural network that evaluates sequences of RGBP images and outputs a real number estimating the approximate people count.

An appropriate training set is compiled by adapting the strategy proposed in \cite{retailnet}. In that work, the diversity of the training set was prioritized by collecting RGB videos throughout different times and days of the week from a surveillance camera positioned in a real shoe retail store. The gathered RGB videos were afterwards structured into a set of individual RGBP images.

The annotation step was carried out by a human trainer using a publicly available semi-automatic annotation tool developed for that purpose\footnote{https://ic.ufal.br/professor/thales/retailnet/}.
As manually labeling thousands of images would be an extremely exhaustive process, the authors proposed an efficient semi-automatic visual annotation tool, whose interface allows the user to visually count the number of people. Temporal coherence was exploited by semi-automatically annotating each collected video. After manually annotating the first frame of a video, the video starts playing. Then, the annotator is expected to increase or decrease by one unit the people count of the frame being displayed since only small variations of the people count is expected over time, \emph{i.e.} when an individual enters or leaves the visible area of the store. Consequently, the training time is expected to be in the order of the duration of the videos.

Differently from \cite{retailnet}, we consider each training sample to be a sequence of $T$ continuous RGBP images in time, \emph{i.e} a small RGBP video labeled with the people count of the last image of the sequence. Thus, the aforementioned annotation process can still be utilized here before compiling the training set of annotated sequences.

After training the regression model, videos acquired in real-time from the camera may be evaluated in a similar fashion: the last recorded frames are used to detect the foreground and compose an RGBP video with the most recent images, which is then given as input data to the neural network. Finally, an accurate real-time prediction of the people count is given as output.

\subsection{LRCN-RetailNet architecture} \label{sec:architecture}
The Long-term Recurrent Convolutional Networks were recently  investigated as an alternative for applications that deal with sequences of images (or videos). The key elements of this architecture are 2d convolutional layers and recurrent layers made of Long Short-Term Memory (LSTM) units. By appropriately combining these layers, this architecture is capable of simultaneously extracting spatial features and learning long-term temporal dependencies. In practice, the LSTM units are sequentially fed by the features extracted from each RGBP image by the convolutional layers. The reasons to combine such architectures are two-fold: first, the people count from subsequent images is expected to present a time dependency; and second, features extract from RGBP images through 2d convolutional layers have been previously shown to be relevant to count people.

\begin{figure}[!t]
\centering
\includegraphics[width=0.7\linewidth]{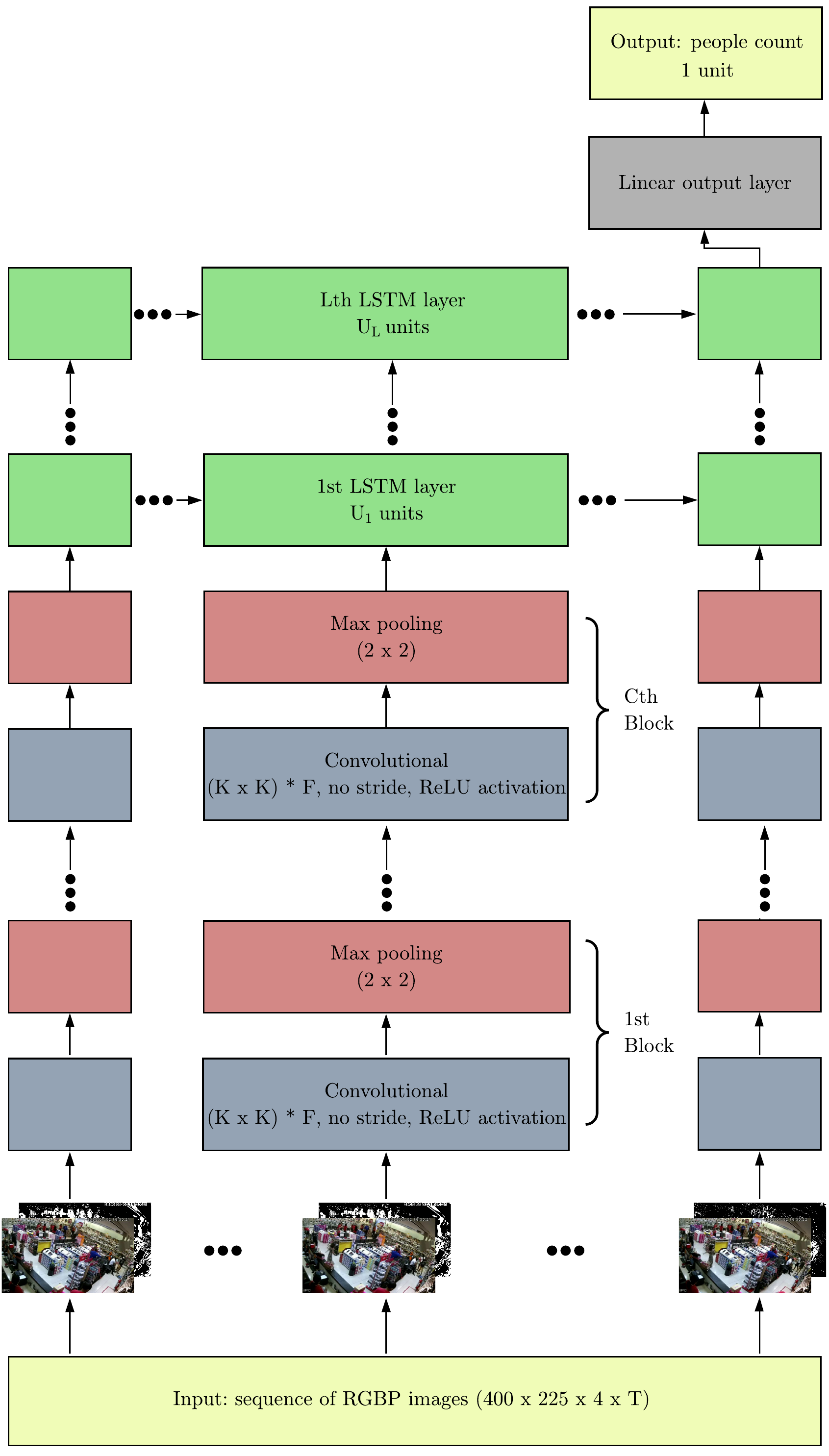}
\caption{LRCN-RetailNet architecture: each image of a sequence of continuous RGBP images in time is given as input to convolutional blocks composed of 2d convolutional layers (each one applying $F$ filters of size $K \times K$) and $2 \times 2$ max pooling layers. Then, the flattened features of each image is sequentially given as input to $L$ recursive LSTM layers that precede a linear output layer with a single neuron. The output of the network is a real number, which is then rounded to its nearest integer, representing the estimated people count.}
\label{fig:lrcn-diagram}
\end{figure}

The structure of the LRCNs investigated in this work is shown in the diagram of \figref{lrcn-diagram}. All networks receive, as input, sequences comprised of $T$ RGBP images with fixed spatial resolution of $400 \times 225$ pixels, resulting in a 4d tensor with dimensions $400 \times 225 \times 4 \times T$. Each RGBP image is individually filtered by $C$ convolutional layers composed of $F$ filters with dimensions $K\times K$, and are always followed by a $2\times 2$ max pooling layer, which will halve the input image in both spatial dimensions and highlight its features. For the convolutional layers, we adopt the standard rectified linear (\emph{ReLU}) activation function. Then, after flattening the resulting feature maps, the resulting array is fed into the recurrent part of the network, which is composed of $L$ LSTM layers, each one consisting of an individual number of units ($U_1,\dots,U_L$).

The features extracted from each image are sequentially given as input to the LSTM layers, whose output flows to the following time step (right), if it is not the last time step; and possibly to a subsequent LSTM layer, if it is not the last recurrent layer, or to the linear output layer, if it is the last recurrent layer and in the last time step (up).

The output layer is made up of a single neuron with linear activation since we are considering the people counting task to be a regression problem. We follow the many-to-one paradigm, which means that the output layer is only fed with data after the last image from the sequence being processed by the previous layers. The output of the last layer is the estimated count for that image, which is afterwards rounded to the nearest integer. Consequently, the output layer estimates the people count by considering local color and foreground features extracted from each image of the input video through the convolutional blocks; and sequentially in time to take advantage of the temporal coherence through the recurrent layers.

\subsection{Customers-only training} \label{sec:customersonly}

In this work, we also intended to adapt and test the capability of the proposed approach to counting only the visible customer in the images, since it is much more important to a retailer to analyze the flow of customers that are visiting the store, without taking into account the salespeople. 

We propose a simple adaptation of the entire approach in which only the annotation of the dataset is modified. By considering that salespeople usually wear uniforms with the same color, we hypothesize that both LRCN-RetailNet and the original RetailNet are capable of learning to extract features in such a way to exclude salespeople from the count.

For this purpose, the label of each image in the training set shall be modified to the number of customers only, disregarding visible salespeople. It is worth mentioning that the same efficient annotation tool described in \secref{supervisedlearning} may be employed in this re-annotation process. In \secref{consumer-only} we describe experiments conducted to validate this adaptation in both the original RetailNet architecture~\cite{retailnet} and in the novel LRCN-RetailNet architecture.

\section{Experiments}
\label{sec:results}
In this section we describe experiments thoroughly performed to answer the following relevant research questions:
\begin{enumerate}
    \item What are adequate hyper-parameters values of the LRCN-RetailNet? 
    \item Do LRCN-RetailNet models outperform the single-image RetailNet?
    \item What is the optimum image sequence size to accurately predict people count?
    \item Can transfer learning and fine-tuning the weights of the convolutional layers of the LRCN-RetailNet architectures boost accuracy when compared to a training from scratch?
    \item Can the proposed approach be easily adapted to customers-only counting?
    \item How do the single-image RetailNet and the LRCN-RetailNet compare to a state-of-the-art approach for object detection such as YOLO~\cite{Redmon_2016_CVPR}?
\end{enumerate}

In what follows, we describe the dataset used to validate, evaluate and compare the LRCN-RetailNet (\secref{dataset}). In \secref{transfer-learning}, questions 1 to 4 are investigated. \secref{consumer-only} is focused in question 5, and question 6 is answered in \secref{yolo-comparison}. Computational performance is analyzed in \secref{performance}.

\subsection{Dataset} \label{sec:dataset}
\begin{figure}[!t]
\centering
\includegraphics[width=0.67\linewidth]{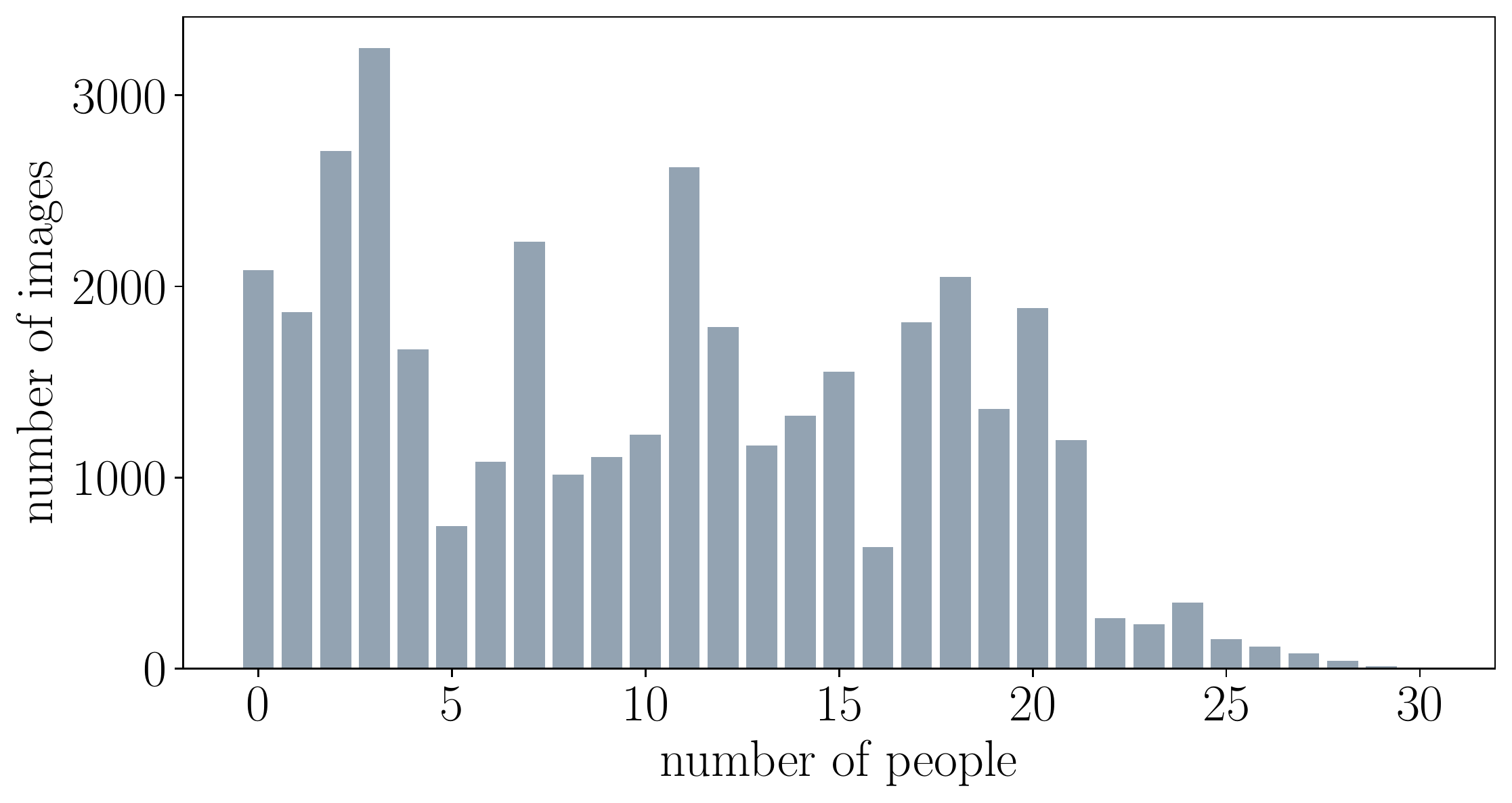}
\caption{Dataset distribution of the number of images per number of people: the same dataset collected and experimented by Nogueira~\etalcite{retailnet} was employed in our experiments. Images showing up to 30 people are included.}
\label{fig:dataset}
\end{figure}

Our experiments were carried out using the same dataset collected to evaluate the original RetailNet work, which is publicly available\footnote{http://www.ic.ufal.br/professor/thales/retailnet/}. The dataset was compiled from many RGB videos that were recorded using a 1-megapixel surveillance camera placed in a real-world retail store. The videos were manually labeled using an efficient annotation tool, described in \cite{retailnet}. As a result, each image in the dataset is labeled with its correct people count. Also, to avoid very similar consecutive frames, only one every five consecutive images were considered. The foreground detection algorithm was applied to extract the P channels from the RGB images, which were then merged to compose RGBP images. The resulting dataset is composed of 37,742 annotated RGBP images showing a wide variety of people counting, ranging from 0 (empty store) to 30 people, as shown in \figref{dataset}.

\subsection{Experiment setup} \label{sec:retailnet}
In all our experiments, we employ the Mean Absolute Error (MAE) as a loss function, and the Adam optimization algorithm~\cite{kingma2014adam} with a learning rate of 0.001 and an early stopping trigger to avoid overfitting.
For each evaluated network, we carried out a cross-validation experiment. The dataset was split by randomly assigning 70\% of the (image) samples for a training set and 30\% for a testing set, in a stratified manner. More specifically, the stratified sampling procedure considered as criterion the people counts of the dataset images to properly balance the sets. 

The accuracy of the network is evaluated on the test set by considering both the MAE, and the error measure
$$
\mathcal{E} = \frac{1}{n}\sum_{\textit{i}=1}^{n}\frac{\left | t_{i} - \text{round}(y_{i}))\right |}{t_{i}},
\label{eq:error}
$$
where \textit{t\textsubscript{i}} is the annotated number of people and \textit{y\textsubscript{i}} is the predicted number of people of the \textit{i}-th example; and round(·) is the function that rounds a number to the nearest integer. Exceptionally, the denominator is set to 1 when \textit{t\textsubscript{i}} = 0. 
To better comprehend and compare the robustness of the models, we also calculate the number of examples that resulted in each absolute error, according to
$$\mathcal{A}(t,y) = \left | t-\text{round}(y)) \right |,$$
where \textit{t} is the ground truth people count and \textit{y} is the network prediction.

We adopt the original RetailNet architecture~\cite{retailnet} as the baseline method for comparisons. More specifically, we compare our results to the best network architecture found in the original work, whose best hyper-parameters values were found to be $C=3, F=8, L=2, U=16$ and $K=5$. It is worth mentioning that here we consider $L$ and $U$ to be the number of dense layers and the number of units per dense layer, instead of the LSTM hyper-parameters defined in \secref{transfer-learning}. 

\begin{figure}[!b]
\centering
\includegraphics[width=0.67\linewidth]{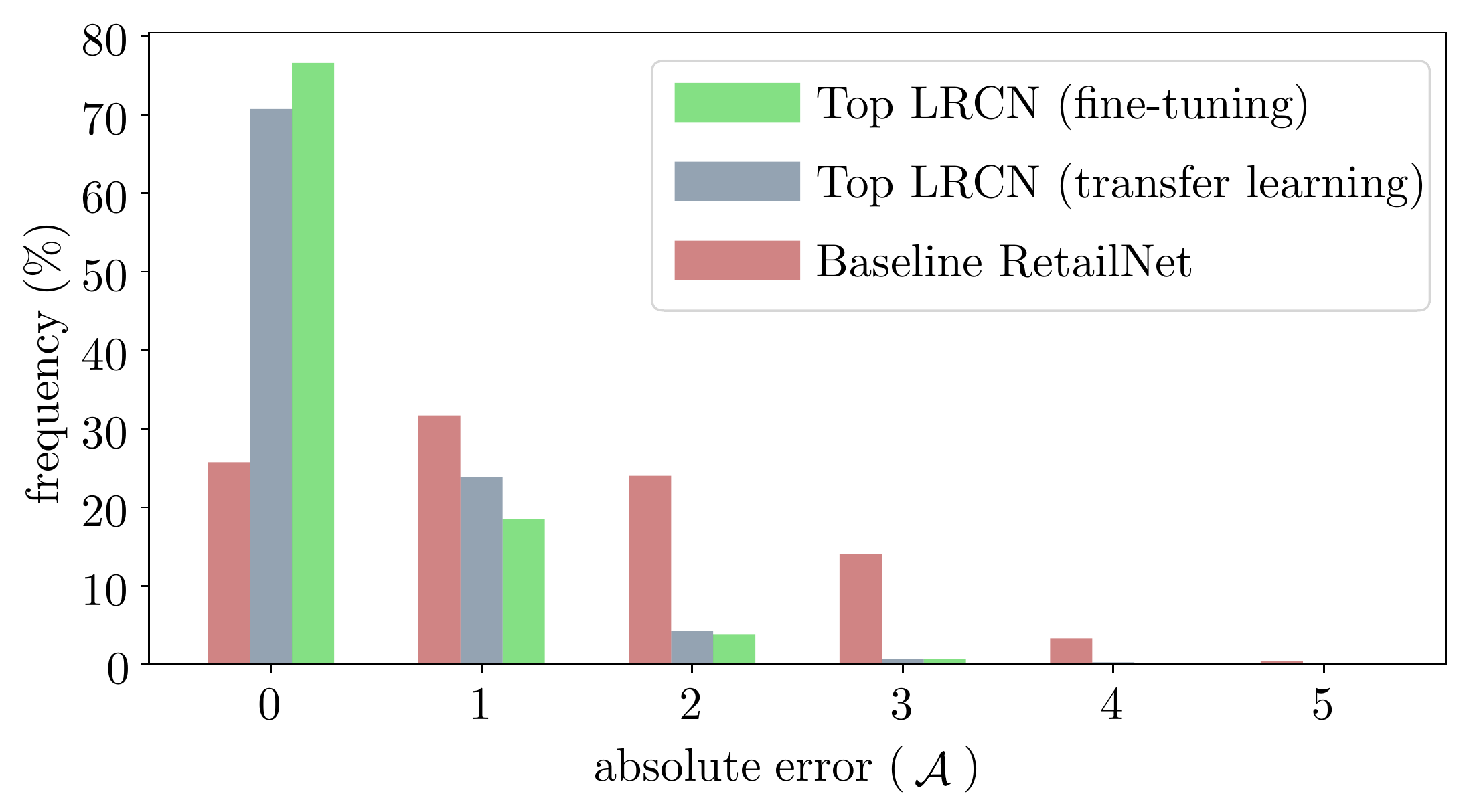}
\caption{Absolute error charts: the horizontal axes denote the different absolute errors $(\mathcal{A})$ achieved by each model on the test set, and the vertical axis denote the relative frequency for each $\mathcal{A}$ value.}
\label{fig:experiments_abs_err}
\end{figure}

The baseline RetailNet achieved $\mathcal{E} = 12.38\%$ and MAE = 1.4 in the aforementioned cross-validation experiments, which are slightly worse results than those found in the original paper ($\mathcal{E} = 10.78\%$ and MAE = 1.232). This difference may be explained by different sampling strategies: the original paper considered a higher proportion of 75\% of the samples for training and stratification was not performed. 
In \figref{experiments_abs_err}, the percentage of test images per absolute error are revealed. Most images show a people count error of up to 2 persons in more than 80\% of the test images. In what follows, we will further refer to this chart to compare to other approaches.

\subsection{Hyper-parameters optimization of the LRCN-RetailNet architecture}
\label{sec:transfer-learning}
The first experiments carried out on the LRCN-RetailNet architecture were focused on optimizing its hyper-parameters and, consequently, validating the entire approach. We initially reshaped the dataset to be comprised of sequences of $T=9$ continuous images in time, where the number of people in the last image was the target value to be estimated by the network.

Since the baseline RetailNet achieved good results, we opted to perform transfer learning by exploiting the weights of the convolutional layers of the best performing model from \secref{retailnet}. Consequently, we fix the convolutional layers hyper-parameters values of all experimented LRCN-RetailNet architectures to $C=3, F=8$ and $K=5$, and set their weights values to the corresponding values from the baseline RetailNet.
The recurrent hyper-parameters $L$ and $U$ were optimized by performing a grid search and experimenting the following variations of hyper-parameters values:
\begin{itemize}
    \item Number of LSTM layers: $L = \{1,2\}$;
    \item Units per LSTM layer: $U = \{250,500,1000\}$.
\end{itemize}
We also applied a dropout of 30\% in each LSTM layer, to avoid overfitting. Following the \secref{retailnet} experiments setup, we employed the Adam optimizer with the same learning rate and the MAE loss. 
The same cross-validation experiment described in the experiments of \secref{retailnet} was conducted. 

\begin{figure}[!b]
\centering
\includegraphics[width=0.8\linewidth]{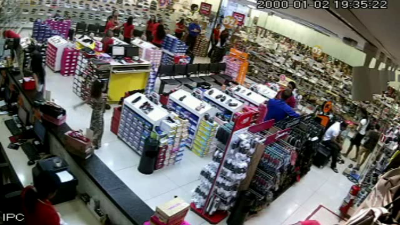}
\caption{Result on a challenging example showing 20 people: a sequence of 9 images (last image shown) was given as input to the best LSTM-RetailNet configuration, that correctly predicted the people count. The single-image RetailNet, by contrast, incorrectly estimated 16 people.}
\label{fig:comparison-image}
\end{figure}

As shown in Table~\ref{table:transfer-learning-results}, all the top configurations achieved significantly smaller MAE and $\mathcal{E}$ than the best RetailNet model. In particular, the best configuration obtained a very good result of $\mathcal{E}$ = 3.59\%, which is almost a quarter of the baseline RetailNet error (12.38\%). 
Since the best result was achieved by the smaller configuration, we further investigated whether an even smaller network would boost the accuracy. However, the results for a single recurrent layer comprised of 125 units did not outperform the top three configurations.
\figref{experiments_abs_err} reveals that the best configuration was capable of predicting more than 70\% of the test images precisely. In \figref{comparison-image}, we show an example in which the best LRCN-RetailNet succeeded in predicting the correct people count in a challenging example with 20 people, while the single-image RetailNet predicted only 16 people, missing by a good margin.
These relevant results also confirms our expectations that, by exploiting a larger amount of data from a sequence of images, and consequently the temporal coherence of the problem, higher accuracy can be achieved by employing appropriate neural network architectures. 

\begin{table}[!t]
\caption{Performance obtained by the top LRCN configurations and the baseline RetailNet model, sorted by $\mathcal{E}$. Best result shown in bold.}
\label{table:transfer-learning-results}
\centering
\begin{tabular}{cc|cc|c}
L & U           & \(\mathcal{E}\) (in \%) & MAE   & params \\
\hline
1 & (250)       & \textbf{3.59\%} & \textbf{0.365} & 9,083,251   \\
1 & (1000)      & 4.19\%     & 0.401 & 39,333,001           \\
1 & (500)       & 4.27\%    & 0.377 & 18,666,501           \\
1 & (125)       & 5.02\%   & 0.498  & 4,479,126           \\
2 & (1000,250)  & 5.46\%   & 0.549 & 40,583,251           \\
2 & (500,250)   & 5.55\%    & 0.589 & 19,417,251           \\
2 & (500,500)   & 5.67\%    & 0.529 & 20,668,501           \\
2 & (1000,500) & 6.49\%   & 0.510 & 42,334,501            \\
2 & (1000,1000) & 6.49\%   & 0.643 & 47,337,001           \\
2 & (250,250)   & 7.01\%    & 0.653 & 9,584,251           \\
\hline
\multicolumn{2}{c|}{RetailNet} & 12.38\% & 1.402 & 145,641
\end{tabular}
\end{table}

Next, we investigated values for the hyper-parameter $T$, which represents the image sequence size given as input to the network. We repeated the same transfer learning procedure of the previous experiments on the best-performing LRCN configuration, but trying to decrease ($T=5$) and increase ($T=12$) the image sequence size. The results shown in Table~ \ref{table:sequence-size} reveal that, in both cases, the resulting $\mathcal{E}$ did not outperform the initial value of $T=9$, which we fix from now on as the proper sequence size.

\begin{table}[!t]
\centering
\caption{Comparison of different image sequence sizes (T) on the top performing LRCN-RetailNet.}
\label{table:sequence-size}
\begin{tabular}{c|ccc}
\multirow{3}{*}{\textbf{error}}& \multicolumn{3}{c}{\textbf{ sequence size (T)}}  \\
& 5 & 9  & 12  \\
\hline
$\mathcal{E}$   & 3.78\% & \textbf{3.59\%} & 6.03\%     \\
MAE & \textbf{0.337} & 0.365 & 0.528
\end{tabular}
\end{table}


Finally, we evaluated distinct training approaches. In addition to the previously described transfer learning strategy, we also experimented:
\begin{itemize}
    \item fine-tuning the best model trained in the previous experiments with a lower learning rate of 0.0001;
    \item training all layers of the best network configuration from scratch.
\end{itemize}
The results shown in Table~\ref{table:training-strategies} reveal that, even though the baseline RetailNet's convolutional layers weights can deliver good performance to the LRCN-RetailNet, such weights can still be improved through a fine optimization, achieving an outstanding accuracy of $\mathcal{E}=2.98\%$, which is the best result in of work. This enhancement can also be observed in the absolute error chart (\figref{experiments_abs_err}): when compared to the previous LRCN-RetailNet model trained with transfer learning, the number of correctly predicted images increases from 70\% to more than 76\%. 

\begin{table}[!b]
\centering
\caption{Comparison of different transfer learning strategies.}
\label{table:training-strategies}
\begin{tabular}{c|ccc}
\multirow{3}{*}{\textbf{error}}& \multicolumn{3}{c}{\textbf{ training strategy}} \\
& from scratch& transfer learning& fine-tuning\\
    \hline
\(\mathcal{E}\)   & 3.96\%  & 3.59\%  & \textbf{2.98\%} \\
MAE & 0.384  & 0.365 & \textbf{0.299}                   
\end{tabular}
\end{table}

\subsection{Experiments on customers-only count}
\label{sec:consumer-only}
We also investigated if the alternative training proposed in \secref{customersonly} is effective to exclude salespeople from the predicted people counts.
Using the same dataset of the previous experiments, we counted the number of customers on each image (disregarding salespeople) using the same annotation tool, and relabeled them. The same stratified cross-validation procedure performed in the experiments of \secrefs{retailnet}{transfer-learning} was conducted here. The modified dataset was used to retrain both the baseline RetailNet and the best LRCN-RetailNet architecture. We employed the same optimization method of the previous experiments. As in \secref{transfer-learning}, we applied transfer learning from the convolutional layers of the baseline RetailNet (trained specifically for customers-only counting) to the corresponding layers of the LRCN-RetailNet and, after training the other layers, we fine-tuned all layers simultaneously.

The results validated the proposed adaptation by revealing even better results than the general people counting approach. The baseline RetailNet achieved $\mathcal{E} = 3.08\%$ and a MAE of only 0.22. This substantial improvement when compared to the results of \secref{retailnet} may be explained due to a substantial number of missed salespeople in the first experiment. Here, they are not taken into account, and thus do not contribute to increasing the errors.

\begin{figure}[!b]
\centering
\includegraphics[width=0.67\linewidth]{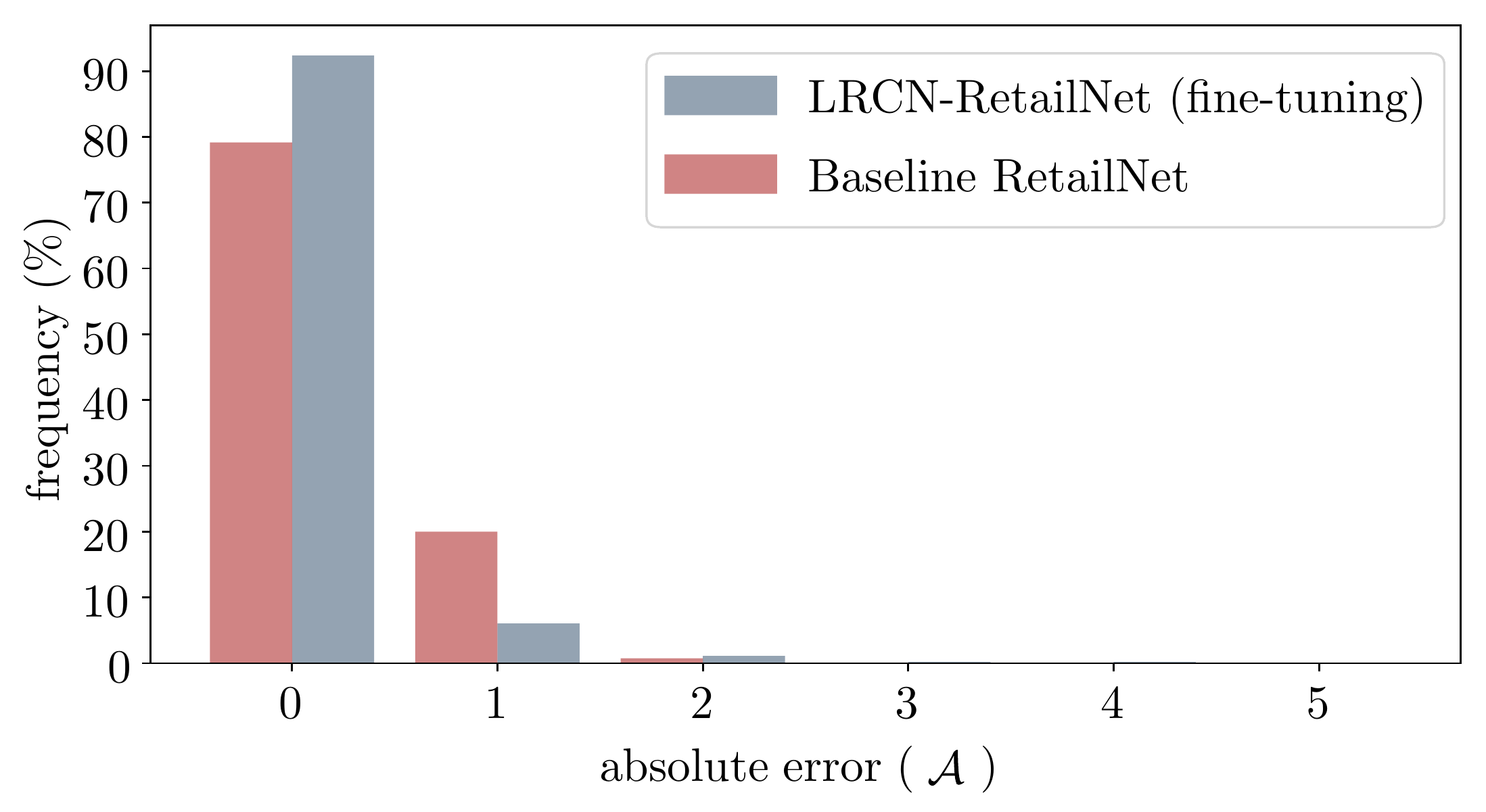}
\caption{Percentage of test images per absolute error: RetailNet achieved a correct estimation of customers count in 79.17\% of test cases while LRCN-RetailNet achieved a correct estimation in 92.37\%.}
\label{fig:con_only_abs_err}
\end{figure}

The LRCN-RetailNet experiments resulted in $\mathcal{E} = 2.33\%$ and an MAE of $0.10$, which is also better than the fine-tuning results found in Table~\ref{table:training-strategies}. It is also worth noting that, once again, the LRCN-RetailNet outperformed the baseline RetailNet. \figref{con_only_abs_err} reveals a high number of accurately predicted examples, achieving more than 92\% for the LRCN-RetailNet. 
These results ratify the proposed adaptation in the labeling process to the task of counting only the customers.

\subsection{Comparison with the object detection approach}
\label{sec:yolo-comparison}
In addition to the comparisons of the LRCN-RetailNet architecture to the baseline RetailNet, we further investigate whether state-of-the-art deep neural networks for object detection can achieve higher people count accuracy. Such approaches try to detect previously trained classes of objects in images, usually providing lists of the detected objects and their respective bounding boxes coordinates.

We compare to the well-known \emph{You Only Look Once} (YOLO, \cite{Redmon_2016_CVPR}). We make use of an open-source implementation of YOLO v3~\cite{yolov3} that is pre-trained in the Common Objects in Context (COCO, \cite{lin2014microsoft}) dataset, with the standard threshold of 25\% for detection confidence. To predict the number of people for each image using YOLO, we simply take the number of objects detected and classified as a person. The same test subset used to assess the LRCN-RetailNet models (\secref{transfer-learning}) was used here.

Results revealed that object detection is not a suitable approach for this task, reaching a poor MAE of 6.241 and $\mathcal{E} = 51.49\%$. A lower confidence threshold of 12.25\% was also tested, but no improvement was found, scoring an MAE of 4.766 and $\mathcal{E} = 52,78\%$. By visually examining some predicted images, we found cases where YOLO missed many visible people (\subfigref{yolo2}), and cases where it detected non-existing ones (\subfigref{yolo1}). 

\tsubimages[!t]{YOLO experiments: many inaccurate predictions were found in our experiments.}{yoloexp}{
\subimage[Example with many false-negatives.]{0.48}{yolo2}%
\subimage[Example with many false-positives.]{0.48}{yolo1}%
}

Our hypothesis for such a poor result from YOLO is twofold. First, YOLO makes use of data from a single image, which makes it difficult even for a human to accomplish the people counting task in some situations. Differently, our approach takes advantage of the temporal coherence of several images in two distinct steps: the background extraction and the recurrent network prediction. Second, YOLO is not trained to cope with very extreme poses and occluded people, which is quite common in the captured images. In contrast, according to \cite{retailnet} the convolutional layers of RetailNet (and consequently LRCN-RetailNet) are capable of learning densities of people per pixel, and thus it is not necessary to effectively recognize each person.
This comparison confirms that our approach is more appropriate to the problem of people counting in retail stores.

\subsection{Performance} \label{sec:performance}
Since the proposed models are intended to be applied as an analysis tool in real-world small retail stores, it is of great importance that they perform well in real-time on affordable computers. Thus, we measured and compared the computational performance of both the baseline RetailNet, which was already shown to achieve real-time prediction, and the best LRCN-RetailNet model. 
Since the networks must be trained for each store and viewpoint, we also evaluate the training times in addition to the prediction times. Furthermore, considering that transfer learning highly decreases training time, we also compared the training time of the LRCN-RetailNet when trained from scratch. The experiments were performed using an Intel Core i5 processor with 4 cores running at 2.20GHz and 8GB of RAM, without using GPU processing. All networks were trained using the Keras Library~\cite{chollet2015keras}, and commonly used Python packages.

The results revealed in Table \ref{table:efficiency} indicate that, as expected, RetailNet has a smaller prediction time, since only a single image is processed. It is also clear that, training the model with transfer learning results in significantly faster training, since the convolutional layers weights are fixed and only recurrent layers are trainable. More importantly, the prediction time for the LRCN-RetailNet allows real-time prediction for people flow analysis, achieving approximately 6 predictions per second. It is worth noting that, since only one every 5 frames are considered (\secref{dataset}) in a 20fps recording, only 4 predictions are required per second, which is practically the prediction time achieved here.

\begin{table}[!t]
\centering
\caption{Experiments on computational performance: the LRCN-RetailNet architectures are shown to be able to achieve real-time prediction and faster training time than the baseline RetailNet when transfer learning (TF) is applied.}
\label{table:efficiency}
\begin{tabular}{c|cc}
\multirow{3}{*}{\textbf{network}}& \multicolumn{2}{c}{\textbf{time per phase}} \\
& training & prediction\\
    \hline
Baseline RetailNet   & 8h45 min  & \textbf{18.53 ms} \\
LRCN-RetailNet without TF  & 12 h & 159.73 ms \\
LRCN-RetailNet with TF   & \textbf{4h15 min}  & 170.12 ms

\end{tabular}
\end{table}

\section{Conclusion}
\label{Conclusion}
We introduced a novel architecture for accurate people counting in retail stores and other indoor environments. As main contributions, we proposed the exploitation of the spatio-temporal coherence of RGB videos acquired from low-cost surveillance cameras through a deep neural network architecture derived from the LRCN architecture. Our experiments revealed significantly higher accuracy when compared to a previous architecture based on single-images and to a state-of-the-art object detection approach, while still achieving real-time performance. The most efficient training strategies and hyper-parameters were  thoroughly investigated. We also showed that a simple adaptation in the labeling task is effective to exclude salespeople wearing working uniforms from the people count. In practice, this result is of great importance for retailers to better analyze the flow of customers. As future work, we intend to research end-to-end neural network architectures that incorporate the foreground detection step in an integrated fashion.

\begin{acknowledgements}
The authors would like to thank CNPq/PIBITI/UFAL for the first author's scholarship and PRMB Com\'ercio e Distribuidora de Cal\c cados LTDA for partially financing this research.
\end{acknowledgements}
\section*{Conflict of interest}
The authors declare that they have no conflict of interest.

\bibliographystyle{spmpsci}      

\bibliography{main.bib}   

\end{document}